%% file: deepfog.tex
\documentclass[sigconf]{acmart}
\usepackage{booktabs} % For formal tables
\graphicspath{{./figures/}}

\usepackage[caption=false,font=footnotesize]{subfig}

\setcopyright{acmcopyright}
\acmDOI{}

\acmISBN{}

\acmConference[MMSys'18]{ACM Multimedia Systems Conference}{June 2018}{Amsterdam} 
\acmYear{2018}
\copyrightyear{2018}

\acmArticle{4}
\acmPrice{15.00}

\begin{document}
\title{Latency and Throughput Characterization of Convolutional Neural Networks for Mobile Computer Vision} 

\author{Jussi Hanhirova}
\affiliation{%
  \institution{Aalto University}
  \streetaddress{Otakaari 24}
  \city{Espoo}
  \state{Finland}
  \postcode{02150}}
\email{jussi.hanhirova@aalto.fi}

\author{Teemu K\"{a}m\"{a}r\"{a}inen}
\affiliation{%
  \institution{Aalto University}
  \streetaddress{Otakaari 24}
  \city{Espoo}
  \state{Finland}
  \postcode{02150}}
\email{teemu.kamarainen@aalto.fi}

\author{Sipi Sepp\"{a}l\"{a}}
\affiliation{%
  \institution{Aalto University}
  \streetaddress{Otakaari 24}
  \city{Espoo}
  \state{Finland}
  \postcode{02150}}
\email{sipi.seppala@aalto.fi}

\author{Matti Siekkinen}
\affiliation{%
  \institution{Aalto University}
  \streetaddress{Otakaari 24}
  \city{Espoo}
  \state{Finland}
  \postcode{02150}}
\email{matti.siekkinen@aalto.fi}

\author{Vesa Hirvisalo}
\affiliation{%
  \institution{Aalto University}
  \streetaddress{Otakaari 24}
  \city{Espoo}
  \state{Finland}
  \postcode{02150}}
\email{vesa.hirvisalo@aalto.fi}

\author{Antti Yl\"{a}-J\"{a}\"{a}ski}
\affiliation{%
  \institution{Aalto University}
  \streetaddress{Otakaari 24}
  \city{Espoo}
  \state{Finland}
  \postcode{02150}}
\email{antti.yla-jaaski@aalto.fi}

%\sharedaffiliation
%\affaddr{Aalto University}  \\
%\affaddr{Espoo, Finland }   \\

    %
    %
%          \alignauthor Sipi Seppäl\"{a}\\    
%          \email{sipi.seppala@aalto.fi}

%   \numberofauthors{6}
%    \author{
%          \alignauthor Jussi Hanhirova\\      
%          \email{jussi.hanhirova@aalto.fi}
    %
%          \alignauthor Teemu K\"{a}m\"{a}r\"{a}inen\\     
%          \email{teemu.kamarainen@aalto.fi}}
    %
%          \alignauthor Sipi Seppäl\"{a}\\    
%          \email{sipi.seppala@aalto.fi}
    %
%          \alignauthor Matti Siekkinen\\    
%          \email{matti.siekkinen@aalto.fi}
    %
%          \alignauthor Vesa Hirvisalo\\    
%          \email{vesa.hirvisalo@aalto.fi}
    %
%          \alignauthor Antti Yl\"{a}-J\"{a}\"{a}ski\\    
%          \email{antti.yla-jaaski@aalto.fi}
    %
%          \sharedaffiliation
%          \affaddr{Aalto University}  \\
%          \affaddr{Espoo, Finland }   \\
%    }

% Lyhyempi otsikko headeriä varten
\renewcommand{\shorttitle}{Latency and Throughput Characterization of CNNs for Mobile Computer Vision}

\begin{abstract}
We study performance characteristics of convolutional neural networks (CNN) for mobile computer vision systems. CNNs have proven to be a powerful and efficient approach to implement such systems. However, the system performance depends largely on the utilization of hardware accelerators, which are able to speed up the execution of the underlying mathematical operations tremendously through massive parallelism. Our contribution is performance characterization of multiple CNN-based models for object recognition and detection with several different hardware platforms and software frameworks, using both local (on-device) and remote (network-side server) computation. The measurements are conducted using real workloads and real processing platforms. On the platform side, we concentrate especially on TensorFlow and TensorRT. Our measurements include embedded processors found on mobile devices and high-performance processors that can be used on the network side of mobile systems. We show that there exists significant latency--throughput trade-offs but the behavior is very complex. We demonstrate and discuss several factors that affect the performance and yield this complex behavior.
\end{abstract}




\maketitle

\input{intro}

\input{background}

\input{mobile}

\input{remote_recognition}

\input{object_detection}

\input{discussion}

\input{related_work}

\section{Conclusion}
\label{sec:conclusion}

This paper describes our study on latency and throughput characteristics of CNNs for mobile computer vision, which was based on measurements with real workloads and real processing platforms. On the platform side, we concentrated especially on TensorFlow and TensorRT but we also used several types of hardware and related driver software. Our measurements included both embedded processors found on mobile devices and high-performance processors that can be used on the network side of mobile systems.

Our main finding was that CNNs for mobile computer vision have significant latency--throughput trade-offs, but the behavior is very complex. There a number of different factors that affect the performance yielding the complex behavior. This makes development of automatic optimization mechanisms challenging. However, in many cases the system behavior makes it possible to get significant performance improvements by tuning the systems.

The systems supporting CNN-based computer vision are rapidly evolving. Along the development, we see two important directions for future research. On one hand, attention is needed on the performance characteristics to make system tuning easier. On the other hand, automated means of optimization are to make usage of the system feasible for the large body of mobile system programmers.

% Ackeja ei varmaan submissioon, koska double-blind review? Jep, hyvä huomio.
%\begin{acks}
%This work has been supported by the Academy of Finland (grant %number 297892), Nokia, and Technology Industries of Finland %Centennial Foundation.
%\end{acks}

\bibliographystyle{ACM-Reference-Format}
\bibliography{biblio} 

%\appendix

%\input{system}

%\input{profiling}

%\input{adaptive}

%\section{Evaluation}

\end{document}

%% file: intro.tex
\section{Introduction}

Computer vision has many important mobile applications in both the consumer realm, such as mobile Augmented Reality (AR) and intelligent traffic, as well as in the Industrial Internet realm, such as perception and control in robotics and intelligent video surveillance. 
%The applications need to be able to identify, recognize, or detect objects and provide feedback or trigger actions based on the outcome. Hence, the basis for such mobile computer vision is algorithms for object recognition and detection from images and videos.
The basis for computer vision is object recognition and detection from images and videos. Usually these tasks need to complete with low latency either for the sake of good human user experience or because the actions they trigger have latency requirements.

In the recent years, deep learning, particularly in the form of deep Convolutional Neural Networks (CNN), has become the dominant approach for implementing computer vision algorithms\cite{lecun15nature}. CNNs have proven to be a powerful and efficient way to implement, e.g., object detection and video scene recognition. CNNs are mainly used for supervised learning in which the network is trained for a specific task. While training a network requires usually a large amount of time and computational resources, a single inference operation (e.g., a detection) can be performed in just a few milliseconds\cite{redmon16CVPR}. However, the computation time depends largely on the combination of hardware and software used. The availability of GPUs or other accelerators can speed up the execution of the underlying mathematical operations tremendously but the software on top of it must be able to reap these advantages properly.

Despite the recent progress with deep learning on terminal devices (e.g., \cite{lane16deepx} for mobile devices), mobile applications often rely on remote servers with specific computing capabilities for fast and energy efficient inference. Real-time operation \cite{nishihara2017realtime} is required also from the server side. The facilitation of the related cloud computing is undergoing a change~\cite{yi15mobidata} and the change affects also the way software utilizes the cloud capabilities~\cite{maas2017cloud30}.

As for computing platforms, the scenery has become very diverse. In addition to the rich variety of GPUs applicable to CNN computations, a number of specific accelerators have been developed. The scale varies from small low-power devices (e.g., \cite{snpe}) to warehouse scale computing (e.g., \cite{jouppi17tpu}). Meanwhile the CPU development~\cite{lee2010debunk} has continued and many CPUs offer acceleration for CNN computations. The same diversity applies to runtime systems~\cite{nguyen2017notanother}. As a result, the computational behavior and performance of CNNs for computer vision are not yet well understood.

In this paper, we study the computational behavior and performance of object recognition and detection from images using CNNs. Our focus is specifically on the inference and we use trained networks on mobile devices and remote servers with and without hardware acceleration. 
% The following would be better in the conclusions (VHi: I condensed stuff somewhat and put it in the concl + the next PG)
% Our long term objective is to develop a distributed system to serve a potentially large number of CNNs and users so that it self-adapts its behavior in order to strike desired performance and accuracy trade-offs. Such a system will also leverage different computing tiers, such as consolidated cloud that provides high end computing services and edge servers that allow a bit less powerful computing at mobile network edge with ultra low network latency.
Our contribution is performance characterization of mobile CNN-based object recognition and detection with several different hardware platforms and software frameworks, both for local on-device and for remote inference. We mainly focus on system throughput and latency and the trade-offs between them. We also examine the impact of parameterization of the software tools for model inference as well as some characteristics of the models themselves on the performance. 

Our results show that using CNN based models yields complex performance behavior but also that they exhibit characteristics that enable performance tuning of the typical systems. Currently, though, human expertise is required to get the most out of the hardware. We believe that our results provide valuable input for work towards self-adapting distributed systems for mobile computer vision that would not need such manual performance tuning.%<-- TODO. Jussi: taviiko tätä?

The structure of the paper is as follows. Section~\ref{sec:background} introduces the concept of CNNs, presents the tools and frameworks used in the measurements and justifies why latency and system throughput are important metrics for CNN based computer vision. Mobile object detection and recognition is measured in Section~\ref{sec:mobile} both for a mobile phone and an embedded computing device. Remote measurements are divided into remote object recognition measurements (Section~\ref{sec:remote}) and remote object detection measurements (Section~\ref{sec:detection}). We discuss the findings in Section~\ref{sec:discussion} and the related work in Section~\ref{sec:relatedwork} before concluding the paper in Section~\ref{sec:conclusion}. %TODO related work section?

%% file: background.tex
\section{Computer Vision Using Convolutional Neural Networks}
\label{sec:background}

\subsection{CNNs for Object Recognition and Detection}

Convolutional Neural Networks (CNN) are a specific class of neural networks that are often used as deep architectures, which means that the networks contain several so called "hidden" layers (i.e., more layers, increased depth) in addition to the input and output ones. 
CNNs are used to extract features from images which are subsequently used in the recognition or detection tasks. One of their biggest advantages compared to many other computer vision techniques is the absence of tedious manual feature engineering.

\textit{Object recognition}, also sometimes called image classification, is among the most studied tasks commonly done using CNNs. Such networks output labels corresponding to classes of recognized objects along with some confidence levels when given an image as an input. Vast majority of work on object recognition has focused on improving the recognition accuracy for which the state of the art CNNs include Inception\cite{Szegedy_2015_CVPR,pmlr-v37-ioffe15,Szegedy_2016_CVPR}, VGG, and ResNet. However, some work exists also on providing faster inference with slightly reduced accuracy. The prominent example is MobileNets\cite{howard17mobilenets} which is a set of CNNs for different latency vs. accuracy tradeoffs achieved by using two hyperparameters, namely width and resolution multipliers, to control the number of parameters and required multiply-accumulate operations in the resulting CNN. MobileNets are mainly targeted for mobile and embedded computing. 

Unlike the CNNs for object recognition, \textit{object detectors} are able to tell also where in the picture the objects reside by outputting bounding box coordinates in addition to class labels. Detection is a more complex task than recognition but CNN-based object detectors can leverage the same CNNs for feature extraction that are used by recognition models, as explained in detail in~\cite{Huang2017CVPR}. The more traditional types of detectors, such as Faster RCNN\cite{ren17frcnn}, work in two stages: First, a set of bounding box proposals are generated and in the second stage a class and class specific box adjustments are predicted for each proposal. Although accurate, these detectors have been reported to be relatively slow. To overcome that limitation, detectors that work in a "single shot" manner, where bounding box proposals are not generated in a separate stage, have been developed. Examples include SSD\cite{liu16ssd} and Yolo\cite{redmon16CVPR}. An important part of object detectors is the last step where detections that might correspond to the same object are merged, also know as \textit{non-maximum suppression}~\cite{HosangBS17}. 

\subsection{Software Tools and Frameworks}

Convolutional Neural Networks can be implemented with various machine learning frameworks. The frameworks all have specific formats to model the structure of a CNN network. Despite some efforts, a single universal language for describing a CNN model is not yet possible. The frameworks do however offer conversion tools between different formats with varying support. We mostly use TensorFlow to build and benchmark the tested networks. TensorFlow is an open-source software library for dataflow programming used by Google both in research and production. 

\textit{TensorFlow Serving} is a high-performance serving system for machine learning models, particularly for inference. We use it to benchmark both object recognition and object detection models in Sections~\ref{sec:remote} and \ref{sec:detection}. It has out of the box integration for TensorFlow models and is designed for production-ready systems. More specifically we utilize the gRPC model server implementation of the prediction service proto buffer available in TensorFlow Serving. GRPC is an open source remote procedure call (RPC) system using HTTP/2 for transport.

\textit{TensorRT} is a inference model runtime by NVidia \cite{tensorrt}. It is used to optimize and execute inference models on different GPU platforms, from datacenter GPUs to portable embedded systems with GPU acceleration. We use TensorRT to characterize object recognition and detection models on the embedded Jetson TX2 platform as well as on desktop GPUs to contrast the results obtained with the TensorFlow Serving.

\textit{Snapdragon Neural Processing Engine} \cite{snpe} is a software development kit produced by Qualcomm for running neural network inference on their Snapdragon 800 and 600 series chips. The SDK has tools for converting TensorFlow and Caffe machine learning models into its custom Deep Learning Container (DLC) format. Snapdragon NPE can utilize Adreno GPUs for hardware acceleration if OpenCL library is available on the device. We utilize the Snapdragon NPE in Section~\ref{sec:mobile} together with the mobile implementation of TensorFlow.

\subsection{Experiments and Performance Metrics}

We focus on inference with CNN-based models for mobile computer vision. On one hand, we study on-device situations where typically a single model is loaded and initialized at a time on mobile device for a specific application. On the other hand, we study remote inference scenarios in which potentially large number of clients send inference jobs to a server serving one or multiple models. 

In the latter case, we also study scenarios where there are more CNN inference models running than there are available accelerators (GPUs). Such scenarios stress the system and reveal their performance characteristics but call for mechanisms for resource sharing.
These experiment use the Linux process abstraction to access shared GPU resources.
We also take a look at the effect of NVidia Multi-Process Service server (MPS)\cite{mps} to the latency and throughput behavior of concurrent processes.

The most important metrics for a computer vision system are accuracy, inference latency, and system throughput.
We do not focus on accuracy in this paper and refer the reader to~\cite{Huang2017CVPR} for a study of accuracy of object detectors. Considering on-device vs. remote inference, the key differences in end-to-end latencies are the added network latency for the remote case and the additional latency due to model loading and initialization in the on-device case, as mobile devices would only in specific cases always keep the potentially large models in RAM all the time.

A common way to increase system throughput is to introduce \textit{batching}. Batching executes the CNN in parallel for several different input images, which reduces overall processing time by improving the reuse of filter weights in convolution operations. However, batching increases latency because running a batch of images instead of a single one through the CNN takes more time. Hence, one possible throughput optimization strategy is to set an upper bound for latency and increase batch size until that bound is met.

%% file: mobile.tex
\section{Mobile On-Device Object Recognition and Detection}
\label{sec:mobile}

In this section, we present results from experimentation with on-device object recognition and detection using CNNs. We use two mobile platforms: a state of the art Android smartphone and Nvidia Jetson TX2 which is a GPU powered embedded computing device. The smartphone represents a regular consumer use case, while the Jetson represents an IoT use case.

\subsection{Object Recognition on Smartphone}
\label{sec:android_recognition}
On smartphone we focus on object recognition because detection models have special operations that are not well supported by the mobile inference frameworks.

\subsubsection{Experiment setup}

We use a Nokia 8 smartphone equipped with Qualcomm SoC (Snapdragon 835 with 8-core Kryo CPU and Adreno 540 GPU) and Android 8.0 (Oreo). Frameworks chosen for CNN inference are TensorFlow 1.5-rc1 Java API and Qualcomm Snapdragon Neural Processing Engine (NPE) 1.10.1 which has both CPU and GPU runtime modes. At the time of our experimentation, TensorFlow has only CPU runtime available for Android.

Two object recognition models are extracted from TensorFlow-Slim image classification library: Inception V2 and full-width (1.0) MobileNet V1, both with input resolution 224x224. We freeze the models into TensorFlow protobuf format and convert with Snapdragon NPE SDK conversion tool into its special DLC format. The disk space requirements for each format are similar but depend on the model: Inception V2 takes up 45 megabytes and MobileNet 17 megabytes in the static asset files of the Android app.

In our application, images are captured as 480x640 preview-quality JPEG images by Android's Camera2 API with exposure time fixed to 1/60 seconds to ensure consistent supply of input images. They are then preprocessed into input tensors (resize, crop, and pixel normalization) before feeding them to the neural networks.

The performance of different frameworks and models was evaluated with measurement of execution latencies in two different cases: total latency of inference with a single image (Figures ~\ref{fig:android_latency_inc2} and ~\ref{fig:android_latency_mob}), and throughput of continuous inference with repeating camera capture (Figures ~\ref{fig:android_throughput_inc2} and ~\ref{fig:android_throughput_mob}).

\subsubsection{Results}

\begin{figure}[t]
\centering
\includegraphics[width=1\columnwidth]{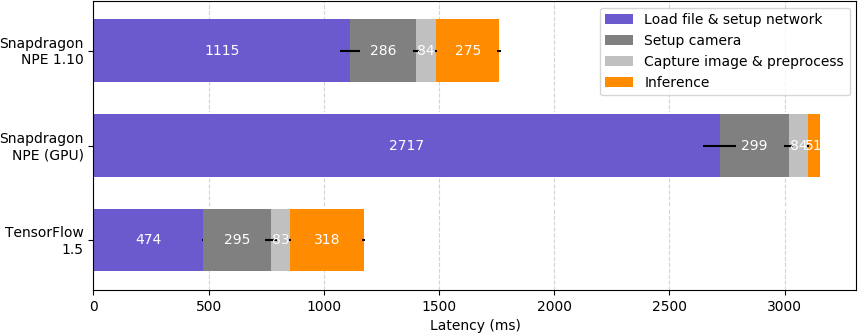}
\caption{Total latency of single image Inception V2 object recognition on Android}
\label{fig:android_latency_inc2}
\end{figure}

\begin{figure}[t]
\centering
\includegraphics[width=1\columnwidth]{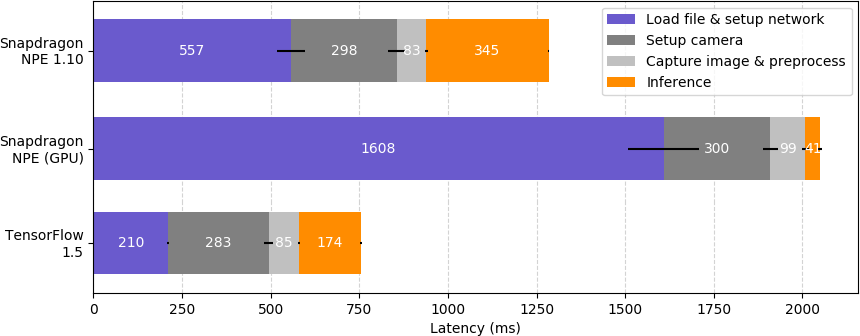}
\caption{Total latency of single image MobileNet object recognition on Android}
\label{fig:android_latency_mob}
\end{figure}

The results in Figures ~\ref{fig:android_latency_inc2} and ~\ref{fig:android_latency_mob} show that for single image object recognition the dominating delay is the model file load and neural network setup, with the actual inference being comparatively fast on all frameworks and both models. Launching the device's camera also takes more time than the actual capture of picture and preparing it for inference. TensorFlow Java API (TensorFlowInferenceInterface) has quite fast network setup, resulting in approximately 750ms total latency for MobileNet where Snapdragon NPE takes more than 1250ms. Interestingly, the NPE's GPU acceleration runtime is useless in this case because the increase in network setup time is longer than the time saved by faster inference when compared to the CPU runtime.

\begin{figure}[t]
\centering
\includegraphics[width=0.9\columnwidth,height=0.6\columnwidth]{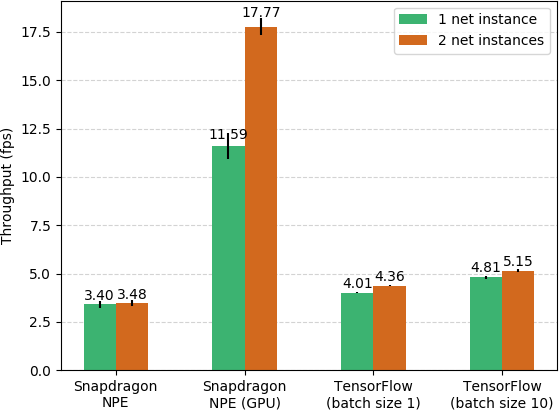}
\caption{Throughput of continuous Inception V2 inference on Android, with one or two neural network instances running simultaneously}
\label{fig:android_throughput_inc2}
\end{figure}

\begin{figure}[t]
\centering
\includegraphics[width=0.9\columnwidth,height=0.6\columnwidth]{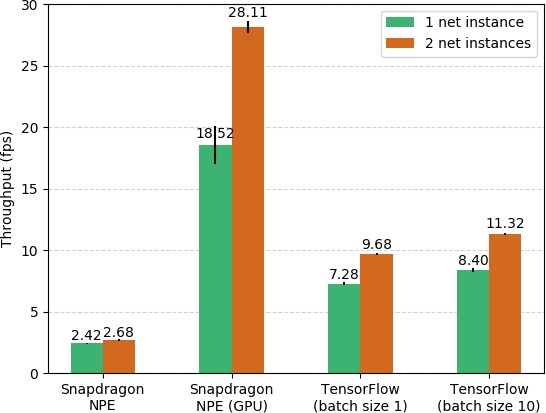}
\caption{Throughput of continuous MobileNet inference on Android, with one or two neural network instances running simultaneously}
\label{fig:android_throughput_mob}
\end{figure}

For the case of continuous inference, Figures ~\ref{fig:android_throughput_inc2} and ~\ref{fig:android_throughput_mob} show frames per second after the initial network and camera setups have already completed. In our application, a frame consists of preprocessing the latest captured image and running inference on a neural network instance, with camera capturing images repeatedly in its own background thread. The results show that throughputs of all frameworks can be increased by loading more than one instance of the neural network API object in their own separate threads, thus providing more concurrency, even when using GPU-acceleration which is already highly parallel. However, using multiple network instances increases the initial setup latency and system memory usage.

\begin{figure}[t]
\centering
\includegraphics[width=\columnwidth]{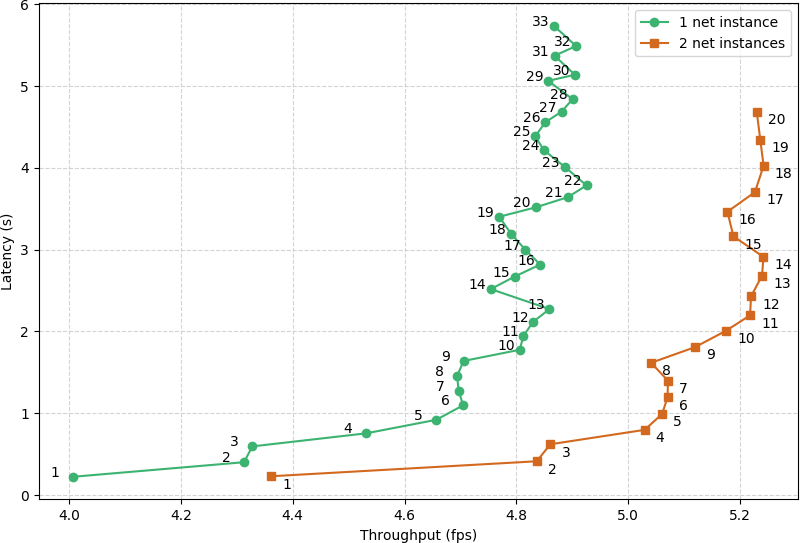}
\caption{TensorFlow 1.5 Inception V2 throughput vs batch latency on Android, with one or two neural network instances running simultaneously taking same number of images per batch. Batch size is varied from 1 to 33.}
\label{fig:android_inc2_tf_batch}
\end{figure}

Figure ~\ref{fig:android_inc2_tf_batch} shows that the throughput of TensorFlow can be further increased by feeding multiple images at a time as a batch. At the time of our experimentation, batching is not available for Snapdragon NPE. In many ways TensorFlow appears to be better optimized for CPU performance than Snapdragon NPE. For example, with batch size of five or more, and two networks running parallel, TensorFlow achieves sub-200ms frame time with Inception and 88ms with Mobilenet, the latter being twice as fast as its 175ms inference-part latency in the single image case. However, in GPU-accelerated mode Snapdragon NPE achieves 56ms frame time with Inception and 36ms with MobileNet which is fast enough for many real-time applications.

\subsection{Object Recognition Using Jetson TX2}
\label{sec:jetson_recognition}

In this section, we characterize the throughput and latency trade-offs of object recognition and detection on the NVIDIA Jetson TX2 embedded computing device. TX2 is an embedded system-on-module (SoM) with dual-core NVIDIA Denver2 + quad-core ARM Cortex-A57, 8GB 128-bit LPDDR4 and integrated 256-core Pascal GPU. The GPU has two streaming multiprocessors.

\subsubsection{Experiment setup}

We measure the throughput and latency of Inception v2 inference model when increasing the image batch size. The experiment is done first with one inference process and then with two concurrently running processes. With one process we adjust the image batch size from 1 to 34 in increments of 1, and with two processes we adjust the batch size from 1 to 16. Input is generated by feeding 300 random batches of images of size 224x224 to the each process. The Inception v2 model is executed on top of the TensorFlow 1.5 runtime. The Inception v2 model is retrieved from the TensorFlow-Slim image classification library.
We also measure the average core utilizations and clock speeds using the \textit{tegrastats} tool during continuous inference with batch size of one image.

During experiments, the device is in the \textit{Max-N} powermode, in which all cores of the TX2 can run at maximum clock speeds but the clock speed is throttled by the Dynamic Voltage and Frequency Scaling (DVFS) of the Jetson TX2 SoM.

\subsubsection{Results}

\begin{figure}[t]
\centering
\includegraphics[width=\columnwidth]{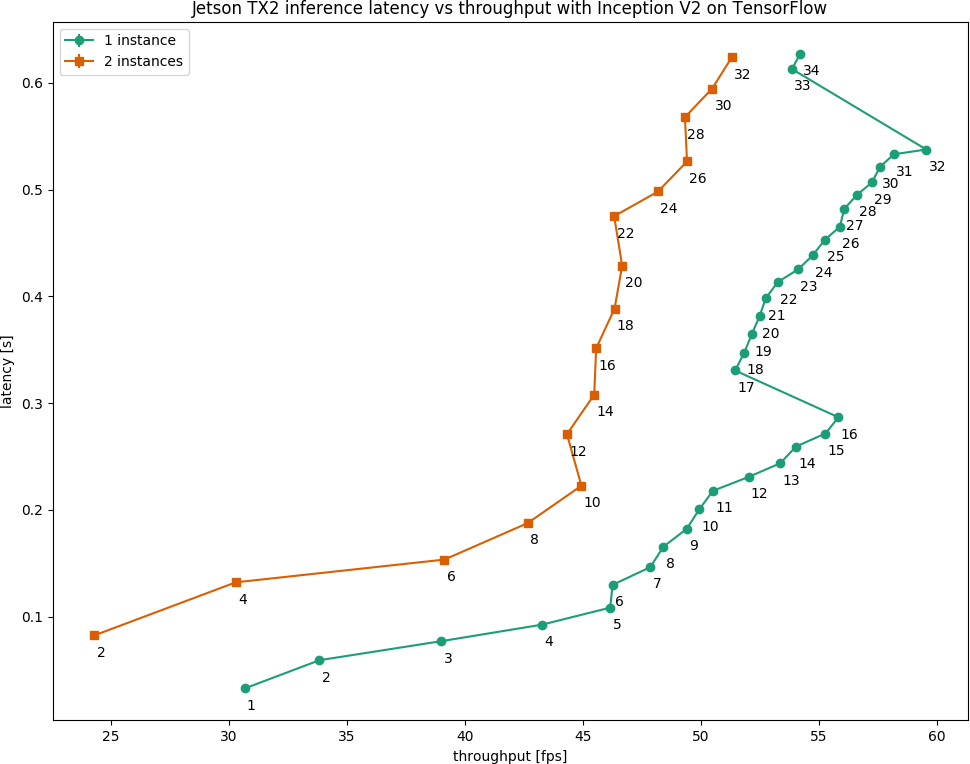}
\caption{Object recognition latency and throughput on Jetson TX2 with Inception V2 model on TensorFlow 1.5. Images are batched together for higher throughput using one and two model instances. Numerical sub-indexing denotes the number of concurrently processed images.}
\label{fig:jetson_lat}
\end{figure}

\begin{figure}[t]
\centering
\includegraphics[width=\columnwidth]{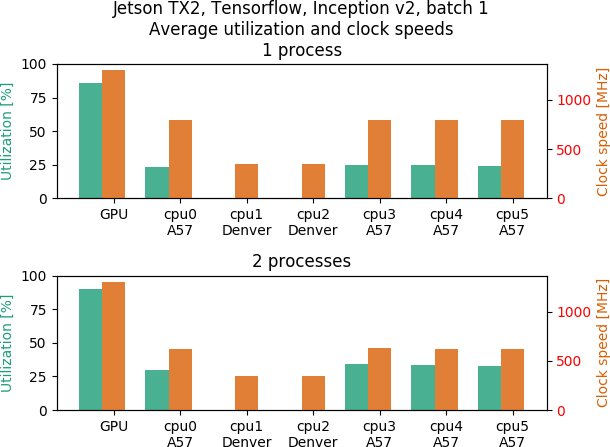}
\caption{Jetson TX2 average utilization and core clock speeds during continuous inference with one and two concurrent Inception v2 model processes.}
\label{fig:jetson_perf_inception_p2}
\end{figure}

Figure~\ref{fig:jetson_lat} shows object recognition latency with TX2 and Figure~\ref{fig:jetson_perf_inception_p2} shows the average core utilization and clock speeds. The average inference latency of one Inception v2 model instance with TensorFlow 1.5 on TX2 is on average 33ms with batch size of one and 540ms with batch size of 32. With two concurrently running model instances the average inference latency is 82ms with batch size of one and 620ms with batch size 16.

In Figure ~\ref{fig:jetson_lat}, the latency-throughput curve of one model instance shows an decline of throughput between batch sizes 16 and 17 and between batch sizes 32 and 33. Observed behavior is due to the TensorFlow runtime choosing different kernels for several inference operations when the batch size increases beyond 16 and 32. For an optimal latency-throughput balance some batch sizes are thus more optimal than others.

In Figure ~\ref{fig:jetson_perf_inception_p2}, the average GPU utilization of one process is 86\%. Both of the Denver cores are idle. The A57 cores are running with average utilization of 25\%. The average GPU utilization of two processes is 90\%. Both of the Denver cores are idle. The A57 cores are running with average utilization of 34\%. With two concurrently running Inception v2 inference model processes the Jetson TX2 has a slightly higher GPU utilization and similarly slightly higher A57 core utilizations. However, at the same time the A57 cores are running on lower clock speeds.

In general, Figure ~\ref{fig:jetson_lat} shows that the execution of two inference model instances concurrently induces execution overhead and the overall latency increases when compared to executing a single model. This is mostly due to context switch overhead of scheduling the two processes as described in ~\cite{Amert2017GPUSO}.
The observed behavior is opposite of the results from the Android experiment presented in Figure~\ref{fig:android_inc2_tf_batch}, where two model instances yield better throughput compared to only one instance.

With both the Android and Jetson the latency-throughput increases are not linearly dependent on batch size. Instead, sometimes an increase of one image in the batch size leads to a better throughput with a minimal increase in latency, and on other other occasions the increase results in both worse throughput and increased latency. Depending on the inference model and the computation platform certain batch sizes are thus more optimal than others. The optimal batch size is difficult to estimate without actual measurements.

\subsection{Object Detection Using Jetson TX2}
\label{sec:object_detection_jetson}
We now turn the attention to object detection models and study the inference performance of three different detectors on the Jetson TX2 embedded computing platform.

\subsubsection{Experiment setup}

The three object detectors used in this experiment are SSD Mobilenet v1 COCO, SSD Inception v2 COCO, and VGG16 FasterRCNN PASCAL VOC. The SSD ones are from the TensorFlow Object Detection library, while the FasterRCNN model comes included in the Jetson TX JetPack 3.2RC SDK. The SSD models are executed on top of the TensorFlow 1.5 runtime using the Python API. The FasterRCNN model uses the TensorRT 3.0 runtime via the C++ API. We feed 500 random jpg images to each model and measure the average throughput.

\subsubsection{Results}

\begin{figure}[t]
\centering
\includegraphics[width=\columnwidth]{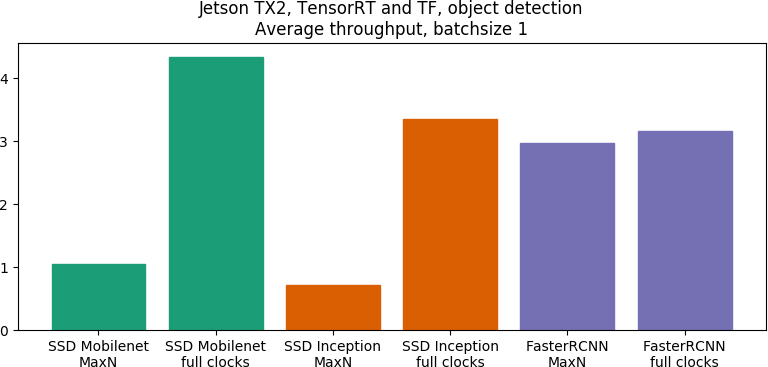}
\caption{Jetson TX2 average inference throughput with TensorFlow SSD inception v2, TensorFlow SSD Mobilenet v1, and TensorRT VGG16 FasterRCNN using batch size of one image. The related core utilizations and clock speeds for the MaxN-powermode are presented in Figure ~\ref{fig:jetson_perf_3}.}

\label{fig:jetson_perf_obj_det}
\end{figure}

Figure ~\ref{fig:jetson_perf_obj_det} shows the average throughput of SSD Mobilenet, SSD inception, and VGG16 FasterRCNN models. 
In the figure MaxN-powermode represents the performance mode of the Jetson TX2. For comparison, the measurement is done also in full clock mode, where DFVS is disabled and all cores run all the time at maximal speed. This represents the theoretical maximum attainable with the system. The average throughputs in MaxN-powermode are 2.7fps for SSD Mobilenet v1, 1.1fps for SSD Inception v2, and 3.2fps for FasterRCNN, while in full clock mode the respective numbers are 4.3fps, 3,3fps and 3.2fps.

\begin{figure}[t]
\centering
\includegraphics[width=\columnwidth]{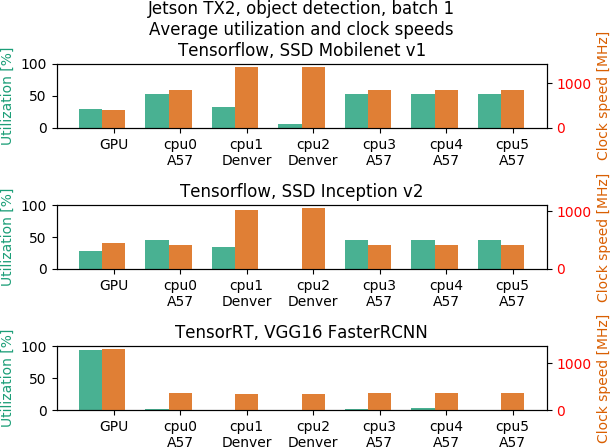}
\caption{Jetson TX2 average core utilizations and clock speeds when running inference using the MaxN-powermode with TensorFlow SSD inception v2, TensorFlow SSD Mobilenet v1, and TensorRT VGG16 FasterRCNN using batch size of one image. The related inference throughputs are presented in Figure ~\ref{fig:jetson_perf_obj_det}.}
\label{fig:jetson_perf_3}
\end{figure}

Figure ~\ref{fig:jetson_perf_3} show the measured average core utilizations and clock speeds in the MaxN-powermode from the throughput measurements in Figure ~\ref{fig:jetson_perf_obj_det}. During continuous inference with the FasterRCNN detector the GPU executes on an average utilization of 95\% with an average clock speed of 1300MHz. The Denver cores are not active and the A57 cores are in practice idle. During inference with the SSD detectors the GPU utilization is 30\% on average. The CPU cores are all active with utilization ranging from 6\% to 52\%. These results reveal that the SSD detectors on TensorFlow runtime are unable to fully utilize the GPU for accelerated inference, whereas the FasterRCNN detector on TensorRT runtime is utilizing the GPU for practically all of the inference operations.

From the models used in this experiment the SSD Mobilenet v1 is generally reported to execute with the highest throughput, and the FasterRCNN model with the lowest throughput~\cite{Huang2017CVPR}. But in practice there are many factors in the interplay of the inference model implementation, runtime support and hardware that dictate the actual performance of a model on a given hardware. The FasterRCNN model used in this experiment runs on top of the TensorRT runtime and is able to use nearly 100\% of the GPU capacity for inference. The SSD models contain operations that have no GPU support on the TensorFlow runtime for Jetson TX2, and cannot thus be executed solely using the GPU, but are instead divided on-the-fly between the GPU and CPUs. With the DVFS enabled the SSD models perform worse than the FasterRCNN model, but when the DVFS is disabled and the Jetson SoM runs at full clock speeds the SSD models perform better than the FasterRCNN model.

Different object detectors or platform configurations would likely lead to yet different behavior than what was observed in our experiments. Finding an optimal system configuration is not a simple task and requires characterization of a large parameter space.

%% file: remote_recognition.tex
\section{Remote Object Recognition}
\label{sec:remote}

The memory requirements together with the long delay caused by model file load and neural network setup limit the attractiveness of on-device inference with mobile devices. Instead, it is often more convenient and even faster to do remote inference, especially when applications need to use several different neural networks. We next study remote object recognition from images using two different standalone server deployments: TensorFlow Serving system and the NVidia TensorRT runtime.

\subsection{TensorFlow Serving}

\subsubsection{Experiment setup}
With TensorFlow Serving, we study the throughput and latency of object recognition using the Inception V2 model on high-end desktop servers. We use two machines, one equipped with an Intel i7 7700K CPU and an Nvidia GTX 1080 GPU (Pascal architecture) and another equipped with an Intel i7 8700K CPU and an Nvidia Titan V GPU (most recent Volta architecture including Tensor Cores). TensorFlow Serving is built using CUDA 9.1 and CUDNN 7.0. We vary both the batch size as well as the number of batching threads which controls the the maximum number of batches processed concurrently.

To measure latency, we modified TensorFlow Serving's gRPC code to measures the time spent during each executed inference call at the server side, hence neglecting the client side and network delays. Throughput with a specific batch size is obtained by dividing batch size by per batch latency. In each experiment, the amount of input images corresponding to the total concurrently processed images, i.e. batch size multiplied by number of batching threads, is sent to Serving in one shot and the next set of input is sent only after all responses from the previous set of input have been received in order not to overload the serving system. In total, 800 images were input in each experiment.

\subsubsection{Results}
\begin{figure}[t]
\centering
\includegraphics[width=\columnwidth]{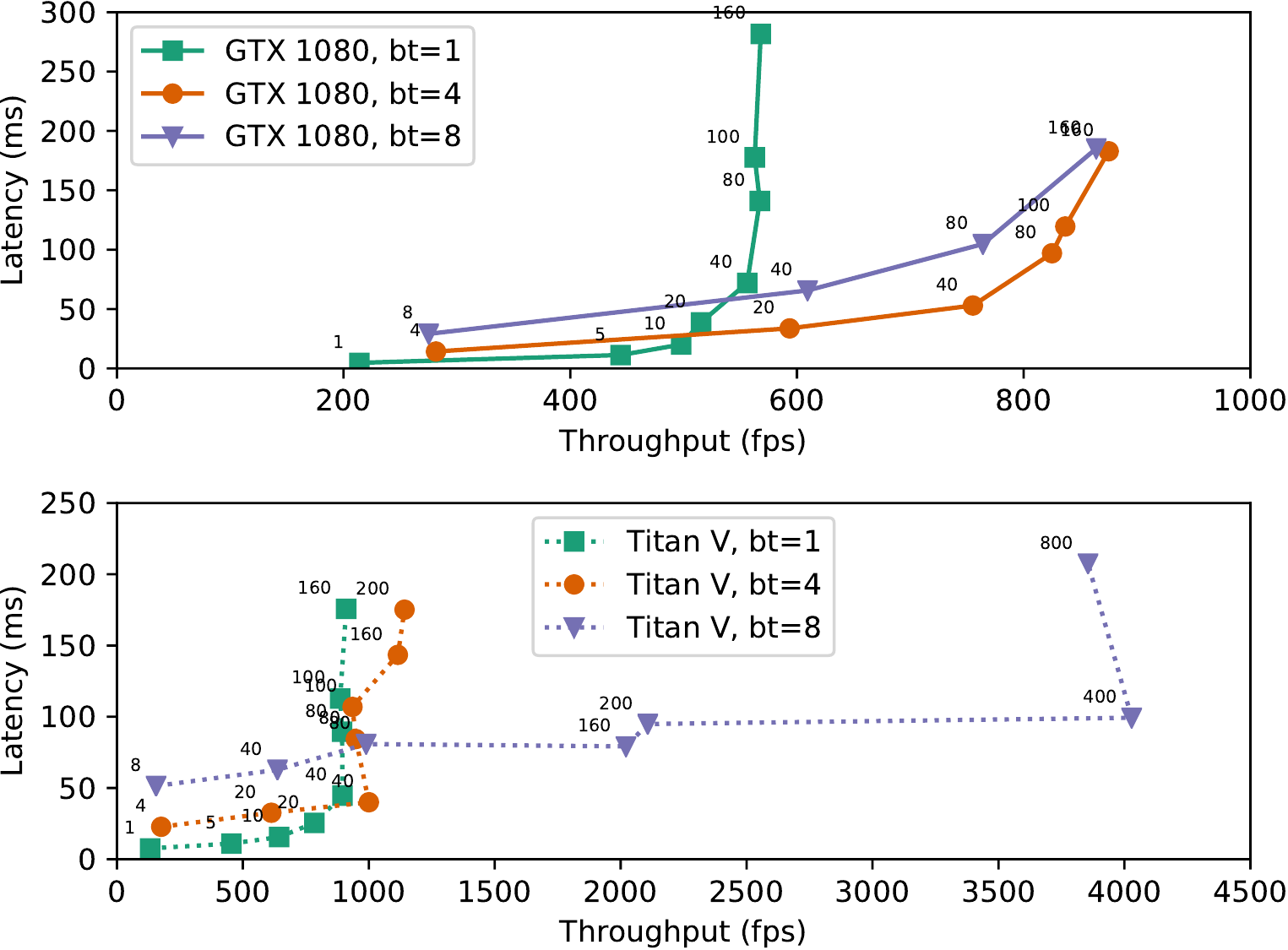}
\caption{Object recognition latency/throughput trade-off when increasing concurrency with Inception V2 using TensorFlow Serving with Intel i7 7700K CPU and Nvidia GTX 1080 GPU (top) compared to Intel i7 8700K and Nvidia Titan V GPU (bottom).}
\label{fig:concurrency_combined_class}
\end{figure}

Figure~\ref{fig:concurrency_combined_class} shows how the inference latency and throghput grow when increasing concurrency. The numbers next to the plots indicate the concurrency which is the batch size multiplied by the number of batching threads. As expected, the lowest latency is achieved with only one concurrent detection. Both overall latency and throughput is optimized using four batching threads with the i7 7700K / GTX 1080 GPU system. The CPU has 4 cores so additional threading does not help beyond this point.

The i7 8700K on the other system has 6 cores which shows in the results. With Titan V the system benefits from batching with significantly higher batch sizes compared to the GTX 1080. With a latency limit of 100 ms, the GTX 1080 has a throughput of 850 images per second while the Titan V can go up to 4000 images per second. The variation of latency was however very high on the Titan V using 8 batching threads (coefficient of variation of latency ranging from 0.2 to 2). This variation can be problematic when designing a production system with latency requirements.

Compared to the mobile scenario in Section~\ref{sec:mobile}, the server deployment can perform roughly 40 concurrent image classifications in the same time as the mobile deployment in its fastest scenario.

\subsection{TensorRT}

\subsubsection{Experiment setup}
Similar to TensorFlow Serving, we examine the latency-throughput behavior with the Inception V2 model using the TensorRT inference optimizer runtime on the same two desktop machines. We use TensorRT 3.0.1 with CUDA 9.0 and CUDNN 7.0. In the experiment, we vary the image batch size from 1 to 160 and perform the measurements using full-precision and half-precision floating point representation for the computations. The GTX 1080 has no acceleration support for half-precision floating point operations (they execute at the same speed as full-precision operations). In this experiment we use random input to the model. The throughput and inference time of for each batch size is measured by averaging the execution time of 100 batched inputs.

\subsubsection{Results}
\begin{figure}[t]
\centering
\includegraphics[width=\columnwidth]{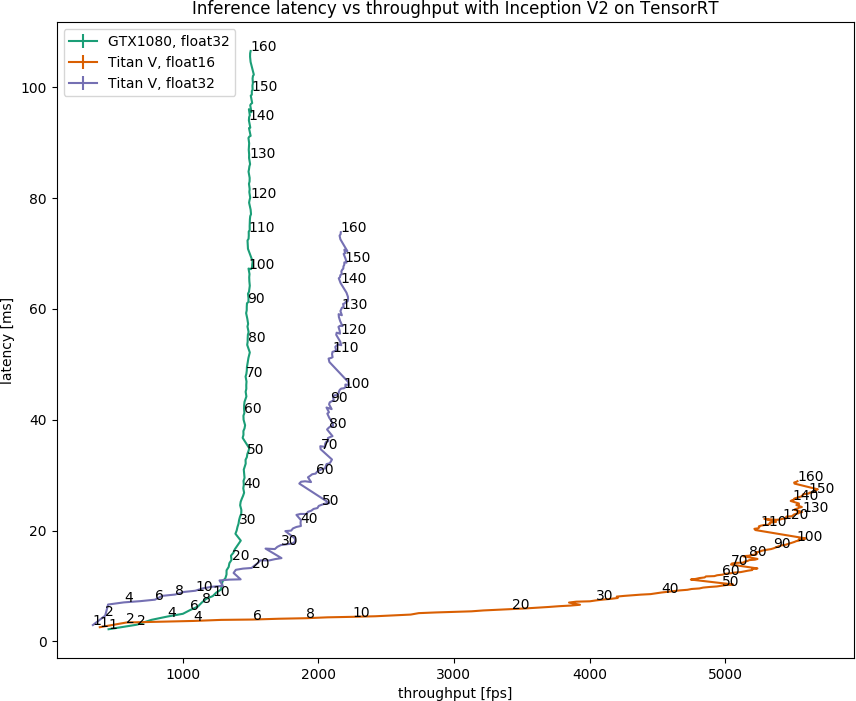}
\caption{Object recognition latency/throughput trade-off when increasing batch size with Inception V2 on TensorRT with GTX 1080 and Titan V GPUs, using full-precision (float32) and half-precision (float16) representation for inference on the Titan V.}
\label{fig:inception_tensorrt}
\end{figure}

Figure ~\ref{fig:inception_tensorrt} presents the results from the experiment. On Titan V with full-floating point representation the latency starts increasing more rapidly after batch size of about 20. The increase in throughput from batch size 50 to 100 is minimal, but the latency grows almost by 100\%. With half-floating point representation throughput increases steadily up to batch size of 100 images. After that, the throughput is not increasing, but the overall latency grows. GTX 1080 has a slightly lower latency with single image inference, but the latency grows fast when batch size is increased.

TensorRT half-floating point support is able to almost triple the throughput of InceptionV2 on Titan V with the higher batch sizes. Also, with half-floats the latency remains relatively low to higher batch sizes as when using full-floating point representation.

Behavior of the latency--throughput relationship is similar in Figures ~\ref{fig:inception_tensorrt} and ~\ref{fig:concurrency_combined_class}. Increases in batch size lead to a better throughput until the system saturates, after which the throughput is not increasing but the latency grows. The actual saturation point depends on many factors, such as the inference model, the underlying hardware, the runtime and its configuration. 

%% file: object_detection.tex
\section{Remote Object Detection}
\label{sec:detection}

We characterize the computational behavior of object detectors also using the two platforms TensorFlow Serving and TensorRT that we used with object recognition. 
In these experiments, we use two recent detectors: 1) SSD meta-architecture combined with the Inception V2 feature extractor in our benchmark model, which has been shown to be fast with a relatively high accuracy, and 2) Faster R-CNN using either the Inception V2 or the VGG16 feature extractor.
We use the model implementations introduced in~\cite{Huang2017CVPR} as they include the full pipeline from image pre-processing to post-processed object detections.

\subsection{TensorFlow Serving}
\label{sec:TF_serving}

\subsubsection{Experiment setup}

The setup and measurement method in the experiments described in this section are the same as in the object recognition experiments with TensorFlow Serving.

\subsubsection{Device placement}

The previous object detection measurements with the Jetson TX2 (Section~\ref{sec:object_detection_jetson}) were performed using TensorFlow's automatic operation device placement. However, the optimal computing device (e.g., CPU, GPU, TPU, Tensor Core) placement of individual operations in a model is an open research question~\cite{mirhoseini17icml}. The operations are still usually assigned to devices by human experts. In this section we experiment with three strategies: 1) Only using CPU, 2) Automatic assignment by the software (TensorFlow) and 3) Manual placement. In automatic mode, TensorFlow maps operations to GPU if the operation has an implementation for it. In the manual strategy we assign pre- and post-processing (mainly non-maximum suppression) stages for the CPU and handle the convolutional network part of the model on the GPU.

\begin{figure}[t]
\centering
\includegraphics[width=\columnwidth]{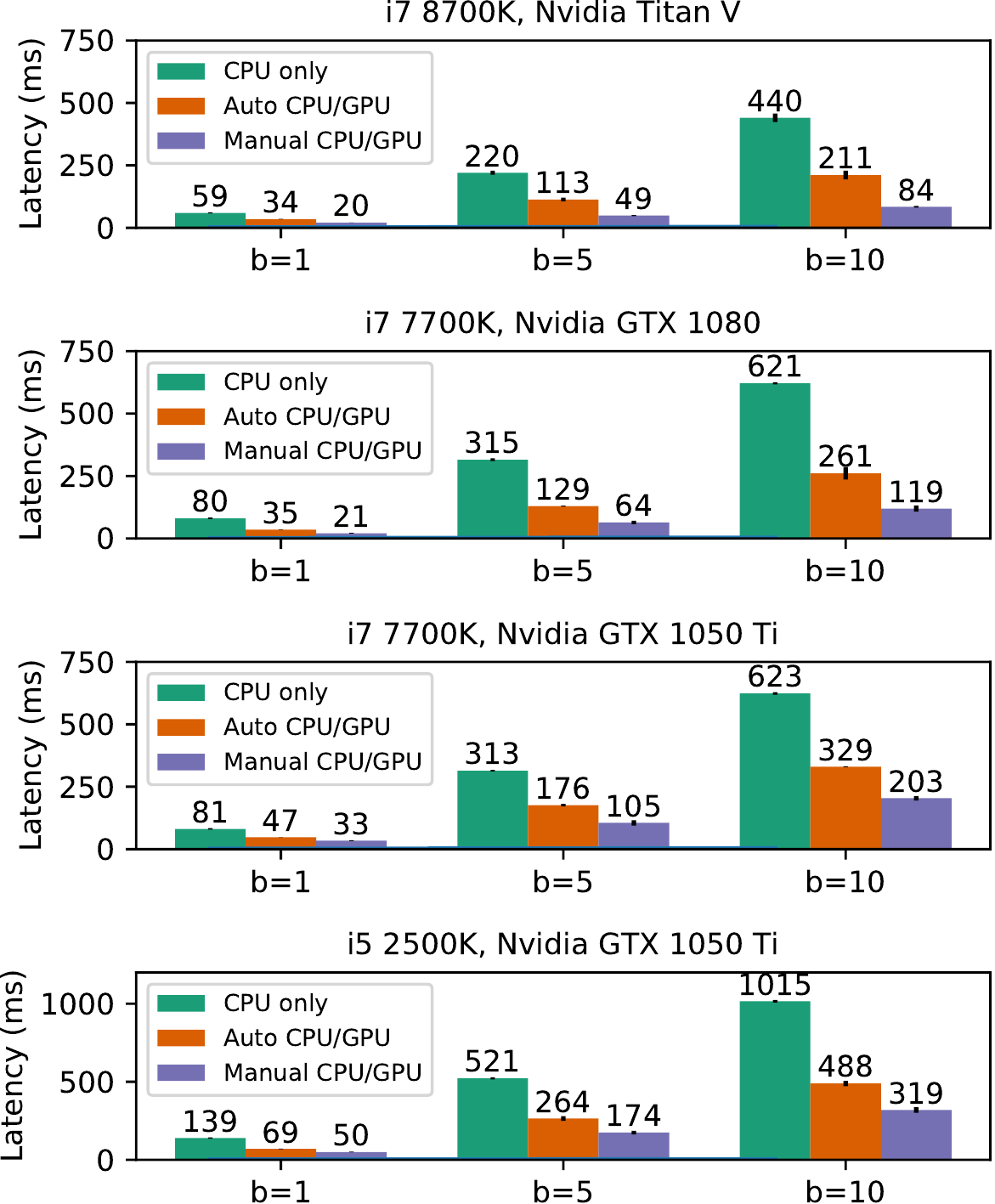}
\caption{Inference latency with SSD Inception V2 using TensorFlow Serving with different CPU/GPU combinations and batch sizes.}
\label{fig:barchart_combined}
\end{figure}

Figure~\ref{fig:barchart_combined} shows how the detection latency rises as we increase the batch size of a single inference using three different GPUs (Nvidia Titan V, Nvidia GTX 1080 and GTX 1050 Ti) and CPUs (Intel i7 8700K, Intel i7 7700K and Intel i5 2500K). The results indicate that TensorFlow by default does not place the device operations optimally as the manual placement strategy outperforms the automatic placement by roughly 20-50\%. Even the high-end CPUs cannot outperform the GPU strategies. We placed the GTX 1050 Ti GPU to a PC equipped with an older CPU (Intel i5-2500K) and a newer CPU (Intel i7-7700K) to see the combined effect of a modern GPU and CPU for performance. The less powerful CPU unsurprisingly runs the model much more slower on its own and also slows down the GPU-powered scenarios by 10-30\%. The Nvidia Titan V GPU shows its power when raising the batch size. With batch size set to one, the Titan V GPU and GTX 1080 have similar latencies.

\subsubsection{Split model}

\begin{figure}[t]
\centering
\includegraphics[width=\columnwidth]{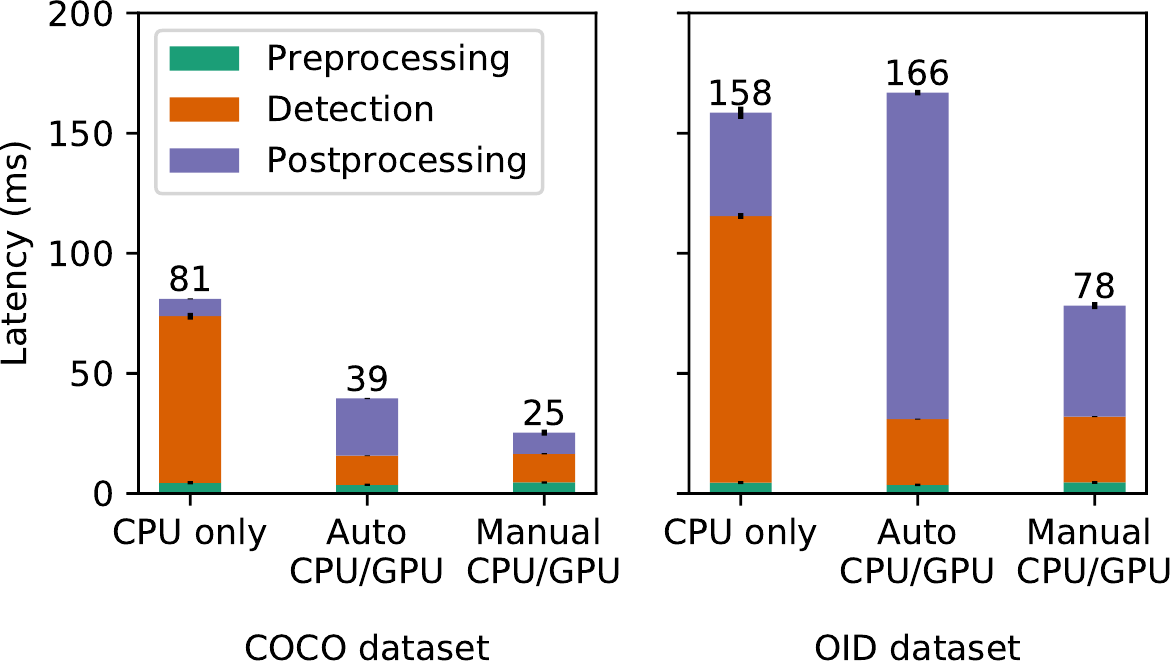}
\caption{Inference latency breakdown with SSD Inception V2 trained with COCO dataset (left) and OID dataset (right) using TensorFlow Serving with Intel i7 7700K CPU and Nvidia GTX 1080 GPU.}
\label{fig:breakdown}
\end{figure}

In order to understand why the manual device placement runs the model faster, we divided the model into three separating the pre- and postprocessing from the rest of the model. We then run the split model parts separately. Figure~\ref{fig:breakdown} shows how the post-processing part runs over two times slower when TensorFlow assigns the operations. Preprocessing is equally fast on CPU and GPU. The rest of the model includes mainly convolutional network operations which run considerably faster when executed on the GPU.

Table~\ref{tab:opcounts} shows the number of operations run on both CPU and GPU in different parts of the model using automatic device placement. The detection part is almost completely run on the GPU with just a couple of memory copies before and after the detection part. This leads to efficient inference when run on the GPU. Preprocessing has almost an equal number of CPU and GPU operations. However there are again only a couple of memory copy operations between the CPU and the GPU which leads to efficient inference when accelerated with a GPU.

\begin{table}[]
\centering
\caption{Device placement and memory copy counts for SSD Inception V2 with automatic placement using TensorFlow. (H=Host, D=Device}
\label{tab:opcounts}
\resizebox{\columnwidth}{!}{%
\begin{tabular}{l|c|c|c|}
\cline{2-4}
                                                 & \multicolumn{1}{l|}{\textbf{Preprocessing}} & \multicolumn{1}{l|}{\textbf{Detection}} & \multicolumn{1}{l|}{\textbf{Postprocessing}} \\ \hline
\multicolumn{1}{|l|}{\textbf{CPU ops}}           & 89                                          & 9                                       & 506                                          \\ \hline
\multicolumn{1}{|l|}{\textbf{GPU op}}            & 106                                         & 433                                     & 5356                                         \\ \hline
\multicolumn{1}{|l|}{\textbf{MemCpy (HtoD)}} & 6                                           & 6                                       & 270                                          \\ \hline
\multicolumn{1}{|l|}{\textbf{MemCpy (DtoH)}} & 6                                           & 8                                       & 734                                          \\ \hline
\multicolumn{1}{|l|}{\textbf{MemCpy (DtoD)}} & 2                                           & 40                                      & 32                                           \\ \hline
\end{tabular}}
\end{table}

Postprocessing part on the other hand shows a high number of memory copies between the CPU and the GPU. This explains the inefficiency of this part of the model when accelerated with a GPU. The memory copies are result of a non-max suppression function which does not have a GPU implementation. This leads to a lot of context changes between the CPU and the GPU as TensorFlow automatically prefers the GPU implementations on other operations.

The results show that it is crucial to separate the parts of the model which can't be effectively accelerated on a GPU for models with non-trivial CPU loads. Otherwise the CPU part might become the bottleneck for the entire model. We further demonstrate this by training the same version of the inference model with a dataset with a larger number of trainable classes as with SSD the number of classes directly affects the postprocessing load (number of detections fed to the non-maximum suppression is a function of number of classes). Figure~\ref{fig:breakdown} also shows how the postprocessing (CPU load) becomes even more dominant part of the overall latency when the model is trained with the Open Images Dataset (546 classes) instead of the COCO data set (90 classes).

We use the manual placement strategy in the next section where we measure the throughput in relation to the inference latency.

\subsubsection{Impact of Concurrency}

\begin{figure}[t]
\centering
\includegraphics[width=\columnwidth]{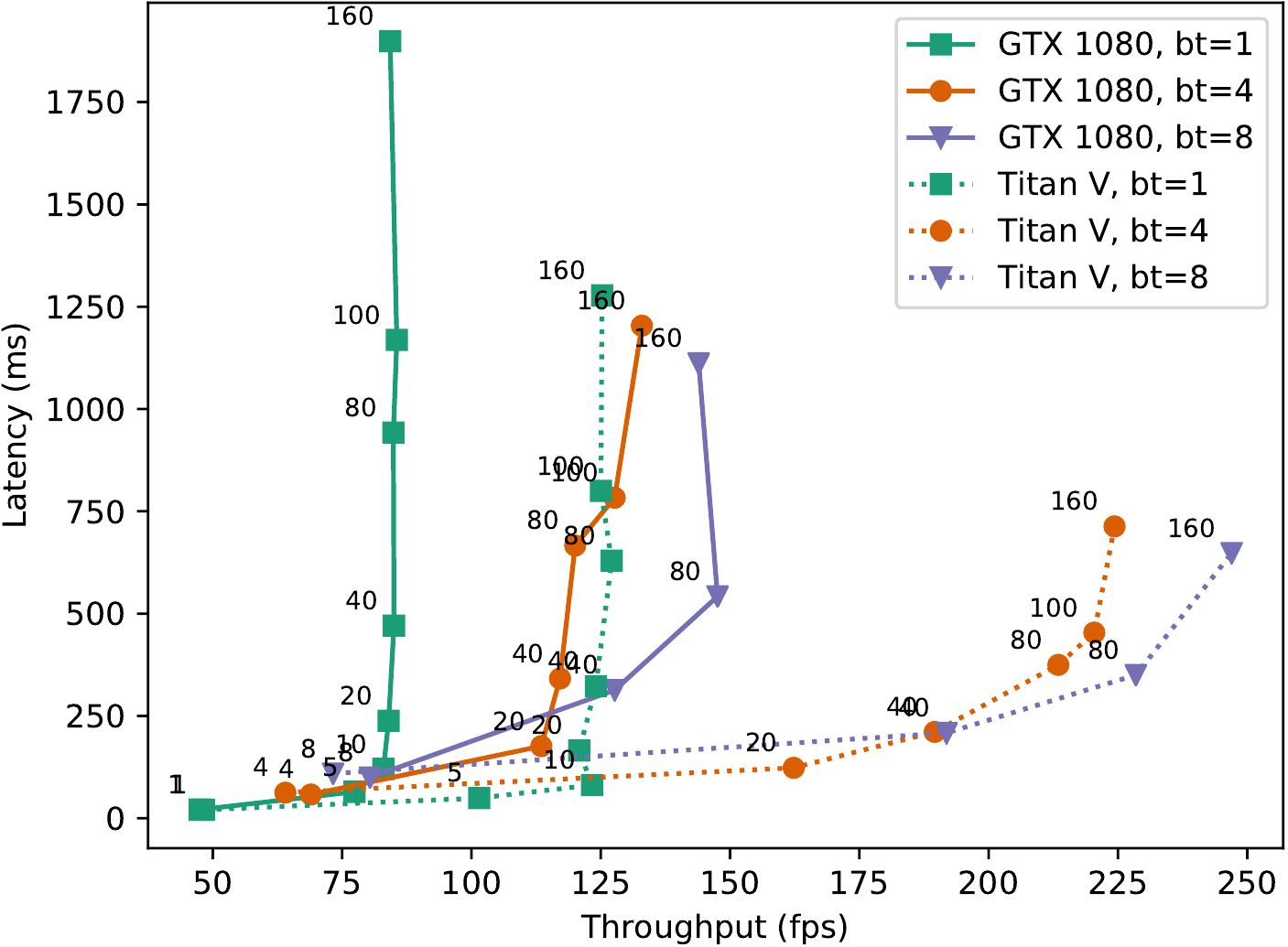}
\caption{Inference latency/throughput trade-off when increasing concurrency with SSD Inception V2 using TensorFlow Serving with either Intel i7 7700K CPU and Nvidia GTX 1080 GPU or Intel i7 8700K CPU with Nvidia Titan V GPU using manual device placement. (bt = batching thread count)}
\label{fig:concurrency_combined}
\end{figure}

The concurrency of the inference can be adjusted in TensorFlow Serving by increasing the batch size and the number of batching threads. The number of concurrent detections can be calculated by multiplying the batch size with the number of batching threads. Figure~\ref{fig:concurrency_combined} demonstrates how the concurrency affects both the latency and throughput with both tested systems. The figure shows that the system throughput can be doubled with the right batching configuration. However, the average latency for object detection grows fast which is a problem for real-time applications.

Similarly to the object recognition benchmarks, the Titan V GPU can handle more concurrent detections than the GTX 1080 with the same latency limit. Depending on the system, a specific batch size must be set to match the latency limit of the application.

\begin{figure}[t]
\centering
\includegraphics[width=\columnwidth]{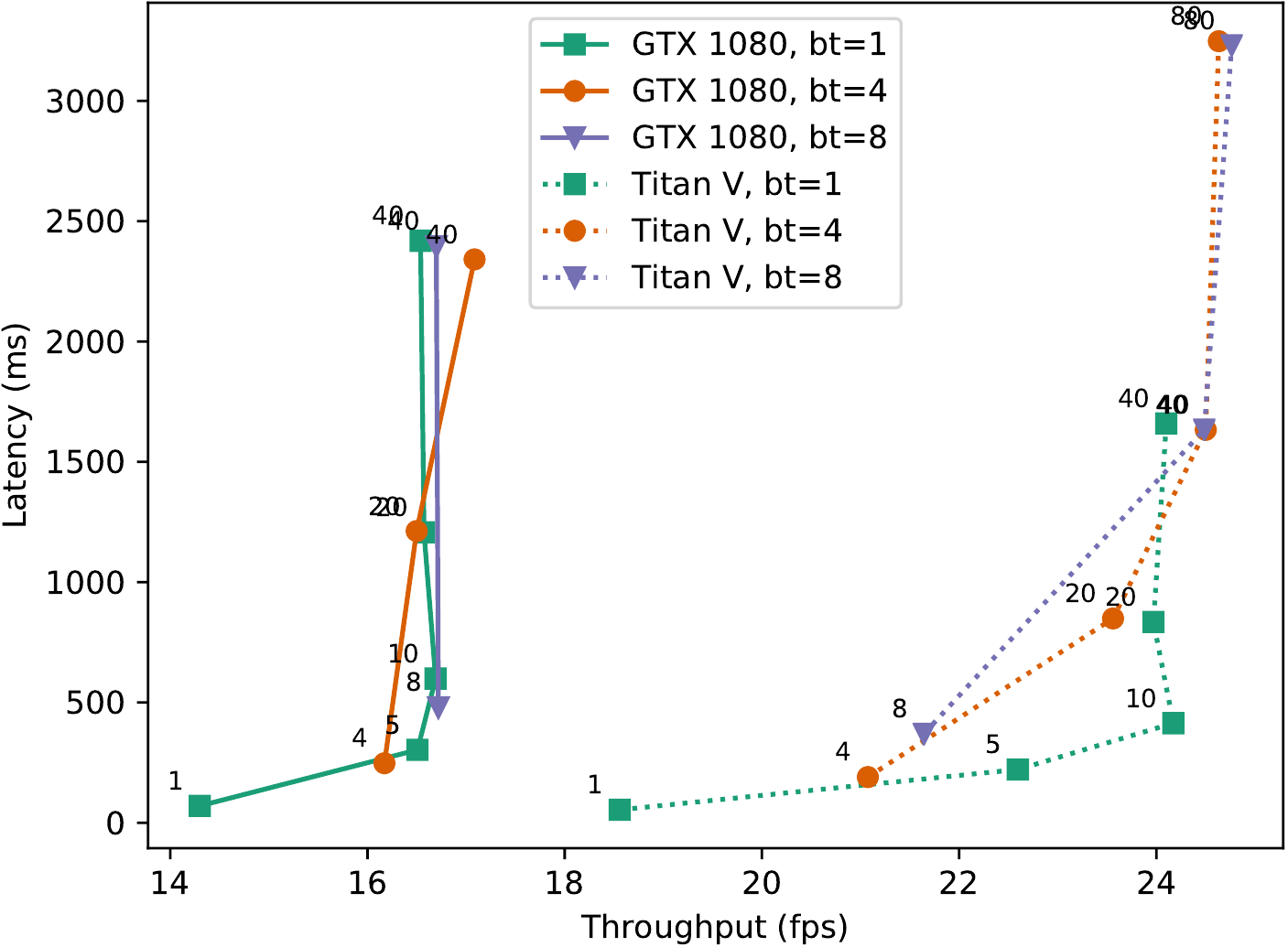}
\caption{Inference latency/throughput trade-off when increasing concurrency with FasterRCNN Inception V2 using TensorFlow Serving with Intel i7 7700K CPU and Nvidia GTX 1080 GPU or i7 8700K with Nvidia Titan V. (bt = batching thread count)}
\label{fig:concurrency_latency_frcnninc}
\end{figure}

Figure~\ref{fig:concurrency_latency_frcnninc} shows the behavior of another object detection model (FasterRCNN InceptionV2) when increasing the concurrency. This model does not have a significant CPU load and increasing the batching thread count does not raise the throughput. The throughput of the model also does not grow much when increasing the batch size.

\subsection{TensorRT}

In the previous section, we used threading together with batching with TensorFlow Serving to increase the concurrency when serving a single model. In this section, we contrast this by performing two experiments with TensorRT to characterize the behavior of model inference when the underlying GPU hardware is shared by different processes that could potentially serve different models. In the first experiment we measure the latency-throughput performance of multiple inference model processes on different GPU hardware. In the second experiment we measure the latency-throughput performance of multiple inference model processes with and without the MPS server. In both experiments we use 1000 random jpg images as input to each process.

\begin{figure}[t]
\centering
\includegraphics[width=\columnwidth]{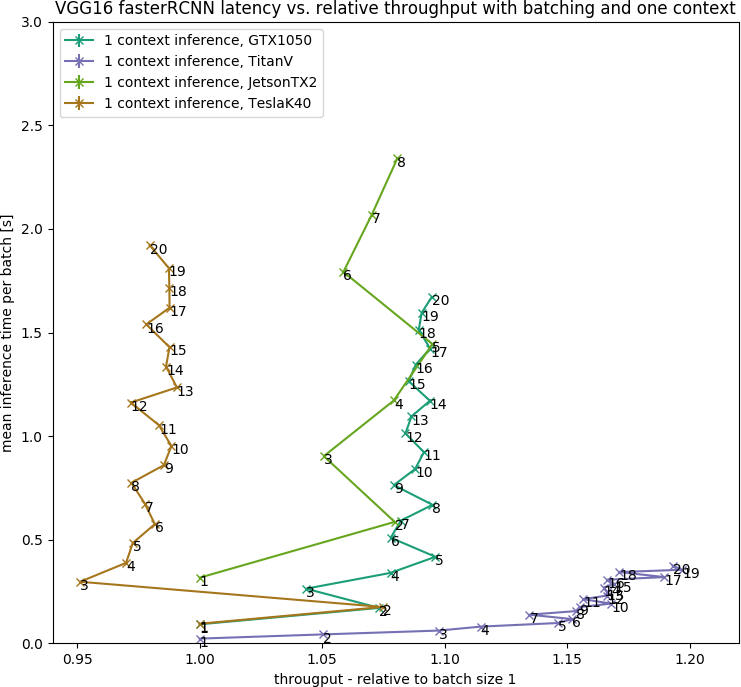}
\caption{Relative throughput speedup with different GPU hardware when batching images using one process context. Batch size is marked on the figure using subindex numbering. Throughput is normalized individually for each GPU by dividing the throughput values by the throughput of batch size one.}
\label{fig:fr_1_rel}
\end{figure}

\subsubsection{Resource sharing}

In this experiment we compare executing one and two processes on different GPUs, namely GTX 1050, Titan V, Jetson TX2, and TeslaK40. Figure ~\ref{fig:fr_1_rel} shows the relationship between the mean inference latency and relative throughput of FasterRCNN detector with different GPUs and varying image batch size. The batch size varies from 1 to 20 for the GTX1050, Titan V, and Tesla K40, and from 1 to 8 for the Jetson TX2.

The newest Titan V is able to achieve the best relative increase in throughput from batching, with a relatively small increase in batch inference latency. Jetson TX2 and GTX 1050 also benefit from batching, but only to a small degree, and their latency increases rapidly with the batch size. The TeslaK40 attains the highest throughput with batch size of two after which the performance degrades.

\begin{figure}[t]
\centering
\includegraphics[width=\columnwidth]{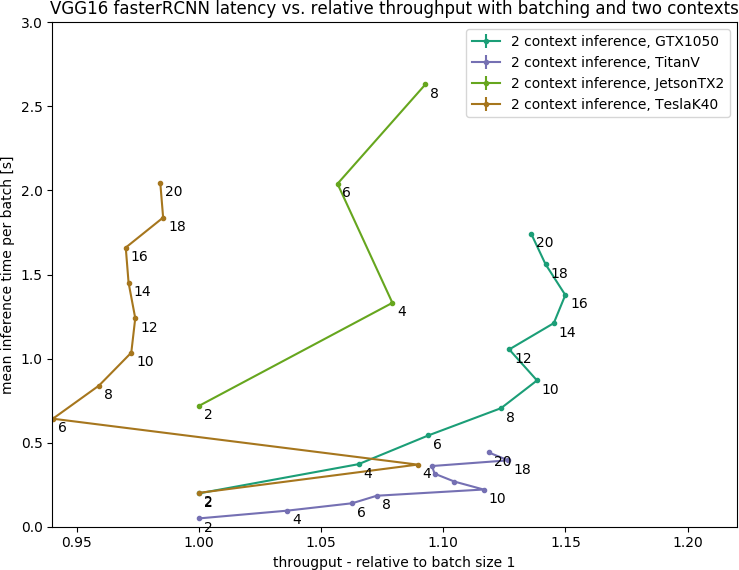}
\caption{Relative throughput speedup with different GPU hardware when batching images using two process contexts. Total number of concurrent images in processing is marked on the figure using subindex numbering. Throughput is normalized individually for each GPU by dividing the throughput values by the throughput of batch size one.}
\label{fig:fr_2_rel}
\end{figure}

In Figure ~\ref{fig:fr_2_rel} the relationship between inference latency and relative throughput of two concurrent FasterRCNN processes is measured using different GPUs by varying the image batch size. The batch size is varied from 1 to 20 for the GTX1050, Titan V, and Tesla K40 GPUs, and from 1 to 4 for the Jetson TX2. With two concurrent processes, the total number of concurrently processed images is double the batch size. The number of concurrently processed images is marked on Figure ~\ref{fig:fr_2_rel} using subindex numbering.

With two concurrently running processes the behavior of relative throughput gain is somewhat similar as with one process. The TeslaK40 achieves the best results with batch size of two per process. The Jetson TX2 gets a small benefit from batching, but the mean inference time increases rapidly. The GTX 1050 gets slightly better throughput speedup from batching with two processes as compared to one process, but the gain is minimal and the latency is also increasing rapidly. The Titan V performs better with one context.

GPU architecture and performance has a remarkable effect on concurrent execution of inference models. It appears that the new TensorRT runtime is not efficient on the older TeslaK40 GPU. The batching support of the fasterRCNN model implementation is not very good, and the relative speedups from batching remain small with all the GPUs used in the experiment.

\subsubsection{MPS}

MPS (Multi-process service) server is acting as a proxy process between user processes and the GPU~\cite{mps}. API requests from different host processes go through MPS and use the GPU as they would come through one process. This way, for example, some kernels that would be serialized can be run concurrently. In this experiment we measure the latency-throughput performance of multiple procesesses running the fasterRCNN model on top of the TensorRT runtime on TeslaK40 GPU. 

\begin{figure}[t]
\centering
\includegraphics[width=\columnwidth]{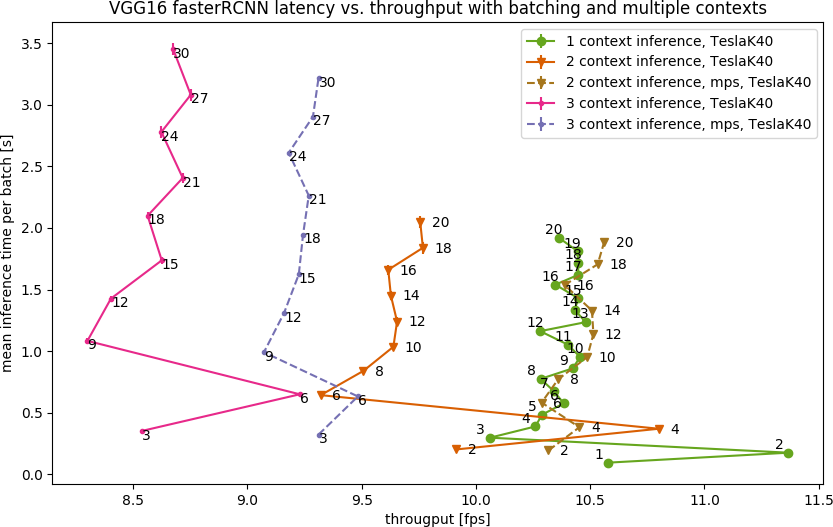}
\caption{Throughput and latency of a single cNN process compared to two concurrent processes. The behavior of two concurrent processes is measured with and without the NVidia multi-process service (MPS).}
\label{fig:fr_tesla}
\end{figure}

Figure ~\ref{fig:fr_tesla} presents the measurement results of batching images for inference using two concurrent processes. Measurement is done with and without the MPS server. As a baseline the behavior of a single process is also included. The measurement is done using the Tesla K40 GPU. Subindex numbering along the measurement points denotes the total number of concurrently processed images. The figure shows that best throughput for multiple contexts is achieved using two execution contexts with batch sizes of two images and without the MPS server. With higher and lower batch sizes using the MPS server with two contexts yields better results. Lowest latency is achieved with two contexts and without batching (if the one procesess baseline is omitted from the comparison).

Figures ~\ref{fig:fr_1_rel}, ~\ref{fig:fr_2_rel} and ~\ref{fig:fr_tesla} show that the behavior of fasterRCNN is not trivial. In Figure ~\ref{fig:fr_1_rel} Titan V GPU achieves throughput speedup with image batching and this induces a relatively small latency overhead. With the GTX1050 and Jetson TX2 the throughput gain via batching is minimal, and already with small batch sizes the maximum throughput is achieved. The TeslaK40 GPU attains the maximum throughput with batch size of two. Larger batch sizes give lower throughput with an increasing latency.

In general, the MPS server helps increase the overall throughput of the system, with the exception of two fasterRCNN processes and batch size of two images, when the behavior is opposite. The optimal number of processes and batch sizes is dependent on the underlying hardware and the inference models and the inference runtimes that are executed in the processes. It seems, that with small batch sizes also the question of whether to use MPS server or not is not a straightforward question to answer. Instead, when one tries to find the optimal latency-throughput configuration for a system, performance profiling of the system is required.

%% file: discussion.tex
\section{Discussion}
\label{sec:discussion}

Machine learning landscape is rapidly evolving. The inference performance of terminal devices (smartphones, IoT devices, etc.) is constantly increasing through specialized hardware, vendor optimized runtimes, and tailored lightweight inference models\footnote{E.g., Android 8.1 introduced Neural Networks API (https://developer.android.com/ndk/guides/neuralnetworks/index.html) which can be used with TensorFlow Lite (https://www.tensorflow.org/mobile/tflite/).}. Similarly, hardware manufacturers are developing more capable CPUs and GPUs, and new types of accelerators such as the TPU and Intel Nervana\footnote{https://ai.intel.com}. 
The still immature state of tools and frameworks poses substantial challenges in model conversion and portability. Execution of same models on different runtimes is not always possible because different runtimes support acceleration of different operations. 

Predicting CNN inference model runtime performance is difficult. The interplay of heterogeneuos computing hardware, optimizing runtimes, software libraries, and (possibly) multiple different inference models sharing the same resources form a multi-dimensional optimization problem.
Usually the optimization space is narrowed by fixing initial parameters such as the GPU architecture or the inference model version with the desired accuracy. These choices already limit many later optimizations. For example, different GPUs behave differently with larger batch sizes or multiple execution contexts (e.g., threads, processes or CUDA contexts). As seen in section~\ref{sec:android_recognition}
Android throughput is increased with two model instances, but with Jetson TX the situation is the opposite.

The accuracy of a model is affected by many things, such as input size or number and complexity of layers. All of these have also an effect on how the model utilizes the underlying resources. In Section~\ref{sec:object_detection_jetson}, different object detection models were characterized on the Jetson TX2 platform. Using the TensorRT runtime which has good support to accelerate operations with the underlying hardware, a computationally more complex model was able to attain better performance than computationally more simple models on a the TensorFlow runtime. But then, by changing the operational mode of the hardware the relative performance of the models changed.

Regarding performance, there are always trade-offs between accuracy, throughput, and latency. While machine learning frameworks implement different tools and methods to automatically optimize their configurations, many parameters still require manual tuning. For example, finding an optimal batch size to maximize throughput with latency constraints requires direct measurements of the inference performance on the execution platform (see Figures~\ref{fig:android_inc2_tf_batch},~\ref{fig:jetson_lat},~\ref{fig:concurrency_combined_class},~\ref{fig:inception_tensorrt},~\ref{fig:concurrency_latency_frcnninc},~\ref{fig:fr_tesla}). Manual placement of operations between GPU and CPU can also significantly improve the execution performance of an inference model (Section~\ref{sec:TF_serving}). Our results indicate that tools and methods available lack performance portability.

Regarding performance characterization, it is difficult to reproduce measurements in full detail because of the many affecting factors. And additionally, in real-world deployments the dynamic and usually unpredictable nature of system input is never the same as in "controlled" measurements.

%% file: related_work.tex
\section{Related work}
\label{sec:relatedwork}

Some CNN frameworks support mobile devices. Recently Qualcomm announced hardware acceleration support for TensorFlow using their latest Snapdragon SoC~\cite{qualcomm}. Some research prototypes that leverage mobile device special purpose processors (e.g., DSP, GPU) also exist~\cite{lane15ubicomp,lane16deepx,lane16mobicase,LatifiOskouei16mm,huynh2017deepmon}.
Other recent research has looked at the computational behavior of CNNs and the impact of the neural network architecture, such as number of layers, depth, etc., on it~\cite{qi17paleo,dong17dnnmark,howard17mobilenets}. Some research has also looked at how to tune the CNNs through model adaptation and optimizations, for instance, in order to tailor them into a specific mobile context and constrained devices~\cite{li16deepcham,han16mobisys,bhattacharya16sensys,huynh2017deepmon}. The custom solutions presented in these papers to utilize  the GPUs and/or DSPs on mobile device are likely to become obsolete with the introduction new tools, such as the Android Neural Network API and SDKs provided by hardware manufacturers (e.g., Qualcomm SNPE). However, their algorithmic optimizations remain valuable. In this paper, we only used the "standard" CNN optimization methods provided by the platforms. Our results also suggest that purely on-device deployment strategies are not desirable in most cases due to memory constraints and long model load and intialization latency.

Concerning CNN inference optimization in distributed systems, Han et al.~\cite{han16mobisys} design a cloud based system for mobile computer vision that executes algorithms either on device or on remote server. Since that work was published, much has changed in short time in terms of the state of the art CNNs for computer vision as well as in terms of software frameworks and hardware acceleration available for neural network inference. That work also focused solely on object recognition, while our results suggest that the computational behaviour of object detection differs substantially from that of recognition. 

Zhang et al.~\cite{zhang17nsdi} studied how to make video analytics systems adaptive in order to manage resources efficiently and with controlled degradation of the quality of analytics. However, the only neural network model used in that work was a DNN classifier and the paper does not specify its architecture or the runtime used to execute the model. The paper also does not present any results with GPUs or any other type of accelerated computing. 

Crankshaw et al.\cite{crankshaw17nsdi} developed Clipper, which is a general-purpose prediction serving system designed for low latency. Clipper is comparable to the TensorFlow Serving.
Mobile on-device inference is not discussed in the paper and the computer vision part includes only object recognition algorithms. One of the most closely related work is the paper by Huang et al.\cite{Huang2017CVPR}. The difference to our work is that they focus only on object detection but also examine the accuracy of the detectors. They also did not study the computational behavior on different computing hardware and runtimes.